\pdfoutput=1
\documentclass{article}
\usepackage{spconf,amsmath,graphicx}

\usepackage{hyperref}
\usepackage{amssymb,amsthm}
\usepackage{xcolor}
\usepackage[ruled,linesnumbered]{algorithm2e}
\usepackage{multirow}
\usepackage{url}
\usepackage{verbatim}
\usepackage{overpic}

\iftrue 
\usepackage{tikz}
\newcommand\copyrighttext{%
\footnotesize Copyright 2022 IEEE. Published in 2022 IEEE International Conference on Image Processing (ICIP), scheduled for 16-19 October 2022 in Bordeaux, France. Personal use of this material is permitted. However, permission to reprint/republish this material for advertising or promotional purposes or for creating new collective works for resale or redistribution to servers or lists, or to reuse any copyrighted component of this work in other works, must be obtained from the IEEE. Contact: Manager, Copyrights and Permissions / IEEE Service Center / 445 Hoes Lane / P.O. Box 1331 / Piscataway, NJ 08855-1331, USA. Telephone: + Intl. 908-562-3966.
}
\newcommand\copyrightnoticeOwn{%
\begin{tikzpicture}[remember picture,overlay]
\node[anchor=north,yshift=-10pt] at (current page.north) {\fbox{\parbox{\dimexpr\textwidth-\fboxsep-\fboxrule\relax}{\copyrighttext}}};
\end{tikzpicture}%
\vspace{-8mm}
}
\fi

\def\bc{{\mathbf{c}}}
\def\bh{{\mathbf{h}}}
\def\bz{{\mathbf{z}}}

\def\bpi{{\boldsymbol{\pi}}}

\title{HALFTONING WITH MULTI-AGENT DEEP REINFORCEMENT LEARNING}
\name{
Haitian Jiang$^{1}$  
\quad Dongliang Xiong$^{1}$
\quad Xiaowen Jiang$^{1}$
\quad Aiguo Yin$^{2}$
\quad Li Ding$^{3}$
\quad Kai Huang$^{1}$
}

\address{
$^{1}$Zhejiang University, Hangzhou, China\\
$^{2}$Pantum Electronics Co.,Ltd., Zhuhai, China\\
$^{3}$Apex Microelectronics Co.,Ltd., Zhuhai, China
}

\begin{document}
\maketitle
\copyrightnoticeOwn 
\begin{abstract}
  Deep neural networks have recently succeeded in digital halftoning using vanilla convolutional layers with high parallelism.
  However, existing deep methods fail to generate halftones with a satisfying blue-noise property and require complex training schemes.
  In this paper, we propose a halftoning method based on multi-agent deep reinforcement learning, called HALFTONERS, which learns a shared policy to generate high-quality halftone images.
  Specifically, we view the decision of each binary pixel value as an action of a virtual agent, whose policy is trained by a low-variance policy gradient.
  Moreover, the blue-noise property is achieved by a novel anisotropy suppressing loss function.
  Experiments show that our halftoning method produces high-quality halftones while staying relatively fast.
\end{abstract}
\section{Introduction}
\label{sec:intro}
Digital halftoning is the technique that converts continuous-tone images into images whose pixel values are limited to only two levels.
Thanks to the low-pass filtering property of our human visual system (HVS), a halftone image can be perceived as its continuous-tone counterpart from a sufficient distance.
According to the dot clustering style, halftoning images can be classified into clustered-dot and dispersed-dot.
Besides, there are two types of halftone textures: periodic and aperiodic \cite{lau2008}.
In this paper, we focus on generating halftones with dispersed-dot and aperiodic features, as this kind of halftone pattern gives the best visual pleasure, which is termed as the ``blue-noise'' property \cite{ulichney1988}.

Three kinds of traditional halftoning techniques are widely used.
Ordered dithering methods \cite{bayer1973,ulichney1993} quantify pixels by comparing the continuous-tone image with a predefined dithering array.
Those methods have high parallelism due to the pixel-wise processing style but sacrifice the halftone quality.
On the other hand, error diffusion methods \cite{floyd1976,ostromoukhov2001,chang2009,liu2015} follow a simple principle: keep the local tone unchanged.
At each step, a pixel is quantized and the quantization error is diffused to adjacent unprocessed pixels.
Even though unpleasant visual artifacts are introduced, error diffusion keeps a good balance between quality and efficiency.
Finally, search-based methods \cite{analoui1992,pang2008} treat halftoning as an optimization problem.
Optimization algorithms such as greedy search or simulated annealing provide pretty good halftone results, while the searching time cost is usually unacceptable on real-time devices.
Considering the halftone quality and the efficiency comprehensively, error diffusion may be the most reasonable method in practical situations, but there is still room for improvement.

\begin{figure}[t]
  \centering
  \includegraphics[width=\linewidth]{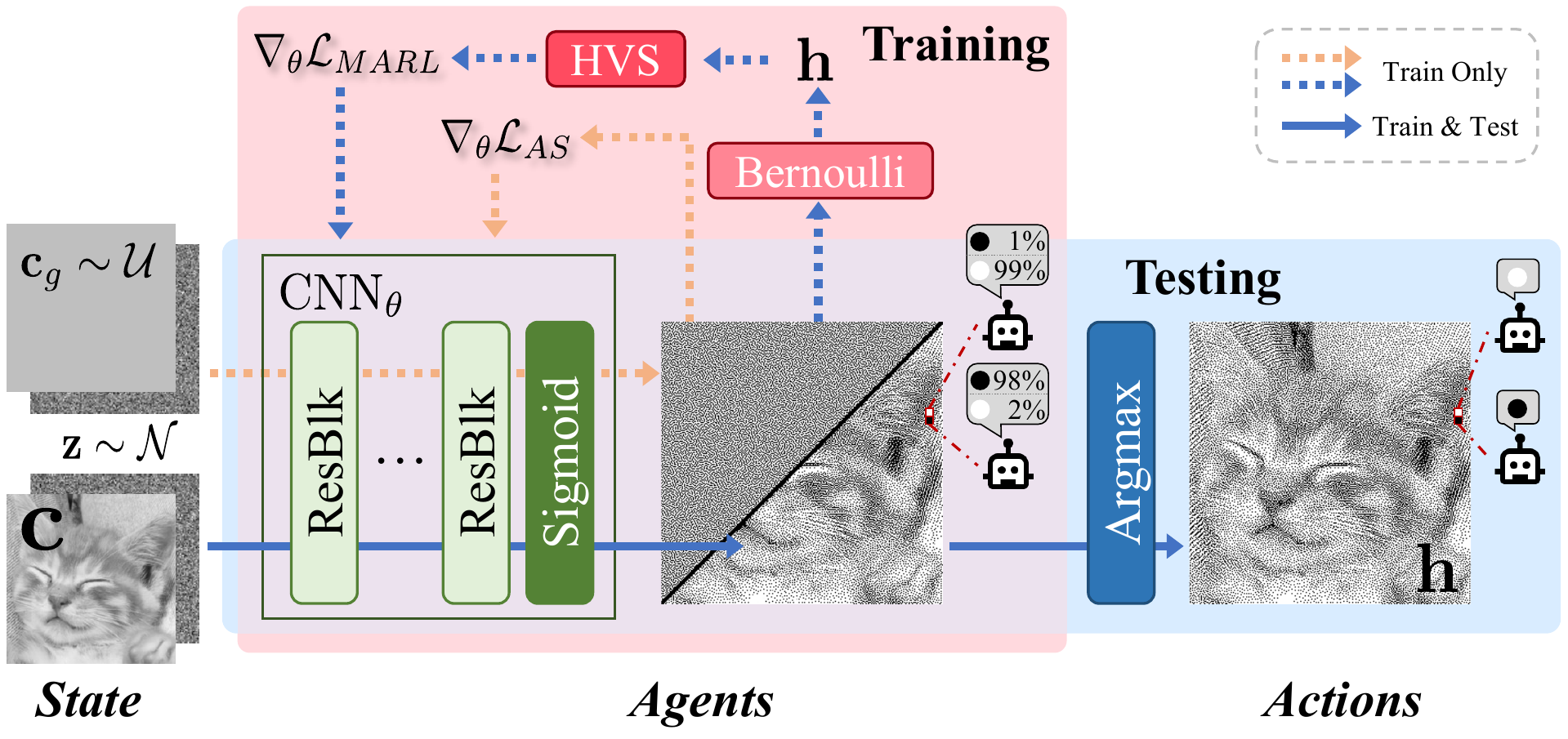}
  \caption{Illustration of HALFTONERS. Each binary pixel in the output halftone image is viewed as an action, which is given by the corresponding virtual agent with a shared CNN policy.}
  \label{fig1}
  \vspace{-4mm}
\end{figure}

Deep neural network (DNN) has recently brought new developments to halftoning.
Kim et al. \cite{kim2018} proposed the idea of rescreening, which dithers an image with another reference halftone image.
They first pointed out that a native convolutional neural network (CNN) can not create position-dependent properties in halftones, and overcame this problem by embedding the position information to the input tensor.
Several works \cite{guo2020,gu2021} viewed halftoning as a unsupervised image translation problem, and tackled it via generative adversarial networks.
However, they all considered only the periodic pattern.
Xia et al. \cite{xia2021} proposed a reversible deep halftoning solution, which introduces a dynamically sampled Gaussian noise map to enable a CNN to synthesize aperiodic halftone patterns.
Although the result halftones make it easier to recover color and details, they fail to achieve a strong blue-noise property, and their two-stage training scheme is complex and unstable.

In this paper, we propose HALFTONERS, a novel deep halftoning method generating halftones with high quality as well as good blue-noise property.
Similar to search-based methods, HALFTONERS tries to maximize a image quality objective metric, but with a parameterized policy.
Specifically, we treat halftoning as a multi-agent reinforcement learning (MARL) problem \cite{foerster2018,gronauer2021}, where pixels in the output halftone image are viewed as the actions chosen by virtual agents, as illustrated in Figure~\ref{fig1}.
The agents' shared policy is parameterized by a CNN, which has high parallelism and can be accelerated by GPUs or NPUs.
Motivated by the counterfactual multi-agent (COMA) method \cite{foerster2018}, we estimate one agent's performance measure by accumulating each action's global reward times the corresponding probability, while sampling other agents' actions.
Thanks to the one-step generating process and the low-dimensional action space, our policy gradient has a lower variance, which leads to a stable training process.
Moreover, to achieve the blue-noise property, we design a loss function which suppresses the anisotropy in the frequency domain.
HALFTONERS is simple to implement, both in terms of training and testing.
Extensive experiments demonstrate the significance of our proposed method in comparison to previous works.
\section{Proposed Method}
\label{sec:method}
In this section, we first present our MARL model for halftoning.
Then an anisotropy suppressing loss is introduced to achieve the blue-noise property.
Finally, we provide a simple and stable training algorithm.

\subsection{MARL Model}
Given a continuous-tone image $\bc \in \mathcal{C}$, a search-based halftoning method generates a halftone image $\bh \in \{0,1\}^N$ by minimizing the visual error $E(\bh,\bc)$, where $E(\cdot)$ is the error metric (e.g. MSE, SSIM \cite{wang2004}) and $N$ is the number of pixels.
Binary pixels $h_a \in \{0,1\}$ compose $\bh$, in which $a \in A \equiv \{1,\ldots,N\}$ identifies the pixel index.
Now we formulate halftoning as a fully cooperative multi-agent reinforcement learning problem (see Figure~\ref{fig1} for an overview).

\noindent \textbf{State.}
The environment state is defined as $\mathbf{s} \doteq \{\bc,\bz\}$, where $\bz$ is the dynamically sampled Gaussian noise map enabling CNNs to dither constant grayness images \cite{xia2021}.

\noindent \textbf{Agents.}
A virtual agent with the shared parameterized policy $\pi$ is constructed at each pixel position, and there are $N$ agents in total.
After observing the state $\mathbf{s}$ and communicating with other agents during the CNN forward pass, each agent gives its probabilities of actions:
\begin{equation}
  P(h_a|\bc,\bz) \doteq \pi_a(h_a|\bc,\bz;\theta),
\end{equation}
where $\theta$ denotes the policy parameters.
The joint policy is defined as $\bpi \doteq [\pi_1,\ldots,\pi_N]$.
We believe that the Gaussian noise map as a prior latent variable offers an image-agnostic dependence source that enables the independent generation of halftone pixels:
\begin{equation}
  \bpi(\bh|\bc,\bz;\theta) = \prod_a \pi_a(h_a|\bc,\bz;\theta).
\end{equation}

\noindent \textbf{Reward.}
The global reward function is defined as:
\begin{equation}
  R(\bh,\bc) \doteq -E(\bh,\bc).
\end{equation}
To keep the notion simple, we leave it implicit that $\pi$ and $R$ are functions of $\bc$, $\bz$, $\theta$, and the gradient is with respect to $\theta$.

\noindent \textbf{Policy Gradient.}
The goal of our MARL model is to maximize the expected global reward:
\begin{equation}
  J(\theta) = \mathbb{E}_{\bc,\bz}\left[\mathbb{E}_{\bh\sim\bpi_\theta}\left[R(\bh,\bc)\right]\right].
\end{equation}
We minimize the negative expected reward with a low-variance policy gradient estimation:
\begin{flalign}
   & \nabla \mathcal{L}_{MARL
  }(\theta) = -\nabla J(\theta) \notag                                                                                                   \\
   & = -\mathbb{E}_{\bc,\bz}\left[\mathbb{E}_{\bh' \sim \bpi}\left[\sum_a \sum_{h_a} \nabla\pi_a(h_a)R(\{h_a,\bh_{-a}'\})\right]\right],
  \label{eqn:pg}
\end{flalign}
where $\bh_{-a} \doteq [h_1,\ldots,h_{a-1},h_{a+1},\ldots,h_N]$ and $\bh'$ denotes the sampled joint actions.
This is motivated by the COMA method \cite{foerster2018}, which assesses one agent's advantage while keeping the other agents' sampled actions fixed.
But what's different is: we directly sum over an agent's all actions, rather than estimating it with its one time sampling like REINFORCE \cite{williams1992}.
It is possible due to the low-dimensional action space (only 2).
The total calculation amount is also acceptable, as an agent only affects a limited area.
Note that the Markov decision process (MDP) in our RL problem has only one step, so there is no need to design another NN to predict the accumulated reward as in actor-critic methods.
The proof of Equation~\eqref{eqn:pg} is given as follows.
\begin{proof}
  \begin{small}
    \begin{equation}
      \nabla \mathcal{L}_{MARL}(\theta) =
      -\nabla J(\theta) =
      -\mathbb{E}_{\bc,\bz}\left[\sum_\bh \nabla\bpi(\bh) R(\bh)\right]
    \end{equation}
    \vspace{-2mm}
    \begin{flalign}
       & \sum_\bh \nabla\bpi(\bh) R(\bh) \notag                                                                                    \\
       & = \sum_{\bh}\sum_a \nabla\pi_a(h_a)\bpi_{-a}(\bh_{-a})R(\{h_a,\bh_{-a}\}) \notag                                          \\
       & = \sum_a \sum_{h_a}\sum_{\bh_{-a}} \nabla\pi_a(h_a)\bpi_{-a}(\bh_{-a})R(\{h_a,\bh_{-a}\})\sum_{h_a'}\pi_a(h_a') \notag    \\
       & = \sum_a \sum_{h_a'} \sum_{\bh_{-a}} \pi_a(h_a')\bpi_{-a}(\bh_{-a}) \sum_{h_a} \nabla\pi_a(h_a)R(\{h_a,\bh_{-a}\}) \notag \\
       & = \sum_{\bh'} \bpi(\bh') \sum_a \sum_{h_a} \nabla\pi_a(h_a)R(\{h_a,\bh_{-a}'\}) \notag                                    \\
       & = \mathbb{E}_{\bh' \sim \bpi}\left[\sum_a \sum_{h_a} \nabla\pi_a(h_a)R(\{h_a,\bh_{-a}'\})\right]
    \end{flalign}
  \end{small}
\end{proof}

\subsection{Anisotropy Suppressing Loss}

\begin{figure}[t]
  \begin{minipage}[b]{0.48\linewidth}
    \centering
    \includegraphics[width=0.7\linewidth]{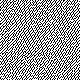}
  \end{minipage}
  \hfill
  \begin{minipage}[b]{0.48\linewidth}
    \centering
    \includegraphics[width=0.7\linewidth]{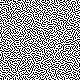}
  \end{minipage}
  \vfill
  \medskip
  \begin{minipage}[b]{0.48\linewidth}
    \centering
    \includegraphics[width=\linewidth]{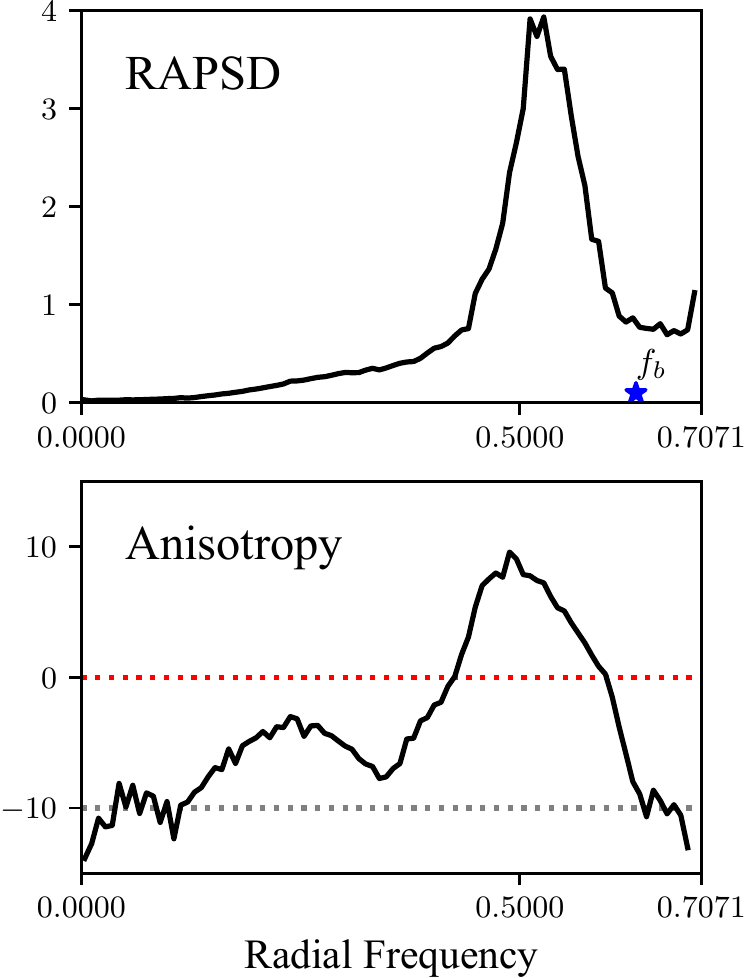}
    (a)
  \end{minipage}
  \hfill
  \begin{minipage}[b]{0.48\linewidth}
    \centering
    \includegraphics[width=\linewidth]{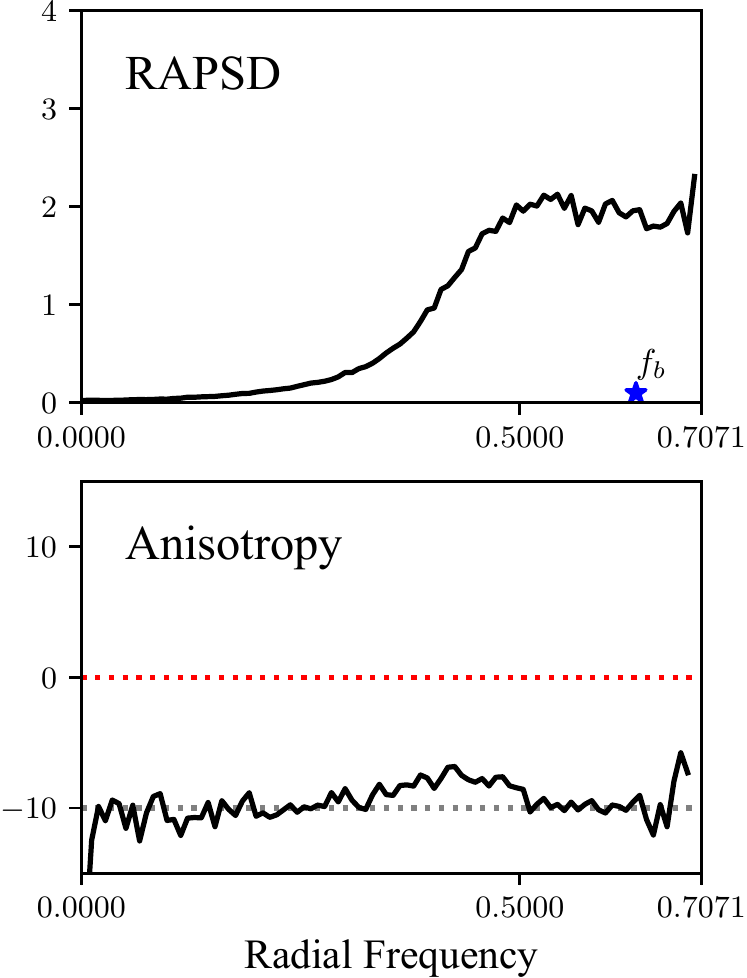}
    (b)
  \end{minipage}

  \caption{Spectral analysis on our model's halftone result of constant-grayness image (grayness=0.6). ``$\color{blue} \bigstar$'' marks the blue-noise principal frequency $f_b$. (a) w/o anisotropy suppressing loss; (b) w/ anisotropy suppressing loss.}
  \label{fig2}
  \vspace{-4mm}
\end{figure}

Although the proposed MARL model enables training a CNN to output non-differentiable discrete images, the desired blue-noise property \cite{ulichney1988} is not explicitly guaranteed.
Even worse, the convolution operation makes it easy to create the stripe artifacts (shown in Figure~\ref{fig2} (a)), as all agents with the same policy tend to enhance the orientation preference.
According to \cite{lau2008}, the blue-noise halftone of a constant grayness image should keep the following spectral characteristics: (a) little low-frequency spectral components; (b) high-frequency spectral region; (c) a peak at the blue-noise principal frequency $f_b$; (d) the anisotropy should be minimized to -10db at all frequency.
Xia et al. \cite{xia2021} proposed to penalize the low frequency.
However, we believe that the excessive anisotropy is the key problem to be solved.

Given the definition of power spectral estimation and radially averaged power spectrum density (RAPSD):
\begin{equation}
  \hat{P}(f) \approx \frac{1}{N} \lvert \text{DFT}(\bh) \rvert^2
  \label{eqn:psd}
\end{equation}
\begin{equation}
  P(f_\rho) = \frac{1}{n(r(f_\rho))} \sum_{f \in r(f_\rho)} \hat{P}(f),
\end{equation}
where $n(r(f_\rho))$ is the number of discrete frequency samples in an annular ring with width $\Delta_\rho = 1$ around $f_\rho$, the anisotropy is defined as:
\begin{equation}
  A(f_\rho) = \frac{1}{n(r(f_\rho))-1} \sum_{f \in r(f_\rho)} \frac{\left(\hat{P}(f)-P(f_\rho)\right)^2}{P^2(f_\rho)}.
  \label{eqn:anis}
\end{equation}
To achieve the blue-noise property, we propose an anisotropy suppressing loss function:
\begin{equation}
  \mathcal{L}_{AS}(\theta) = \mathbb{E}_{\bc,\bz} \left[\left(\hat{P}_\theta(f) - P_\theta(f_\rho)\right)^2\right],
\end{equation}
which minimizes the numerator in Equation~\eqref{eqn:anis}.
We could treat $A(f_\rho)$ as part of the reward in the MARL model, but the power spectrum item would result in dense gradient computations.
Here we take the differentiable probabilities $\bpi(\bh)$ as a proxy for $\bh$ to estimate the power spectrum in Equation~\eqref{eqn:psd}, since it can already capture the anisotropy of $\bh$.
As the spectral analysis only make sense for dithered constant grayness images, we optimize this loss on a batch of uniformly sampled $\bc_g$ in every training iteration following \cite{xia2021}.
As shown in Figure~\ref{fig2} (b), the artifact vanishes as a result of $\mathcal{L}_{AS}$.

\subsection{Training of HALFTONERS}
Now we describe the complete training procedure. To generate halftones with the tone consistency as well as structural similarity, we define the error metric $E$ as:
\begin{equation}
  E(\bh,\bc) = \text{MSE}(G(\bh), G(\bc)) - \omega_s \text{SSIM}(\bh,\bc),
\end{equation}
where $G(\cdot)$ is the HVS low-pass filter and $\omega_s=0.006$.
Following \cite{kim2002} we adopt the Näsänen HVS model \cite{nasanen1984} with a moderate $scale=2000$.
We train the model from scratch optimizing the combined loss function:
\begin{equation}
  \mathcal{L}_{total} = \mathcal{L}_{MARL} + \omega_a \mathcal{L}_{AS},
\end{equation}
where hyper-parameter $\omega_a=0.002$ is set empirically.

In the testing phase, agents give their discrete actions through the simple argmax operation (see Figure~\ref{fig1}).
One may consider a more complex halftone error metric since the assessing computation only exists in the training phase.

\section{Experiments}
\label{sec:experiments}
In this section, we show the implementation details of HALFTONERS and compare both halftone image quality and processing efficiency with prior typical halftoning methods: ordered dithering (VAC \cite{ulichney1993}), error diffusion (Ostromoukhov \cite{ostromoukhov2001}), search-based method (DBS \cite{analoui1992}) and DNN-based method (Xia et al. \cite{xia2021}).
We test all methods on 3,367 images randomly selected from 17,125 images in the VOC2012 dataset \cite{voc2012} following \cite{xia2021}.
To take full advantage of the parallelism in some halftoning methods, we implement them on an NVIDIA GeForce RTX 2080Ti GPU with PyTorch 1.10.1, CUDA 11.3 and cuDNN 8.2.0.
For those serial methods, we implement them in C with GCC 6.3.0 (-O3 optimization) on an Intel Xeon E5-2650v4 (2.2 GHz) CPU.

\noindent
\textbf{CNN Architecture \& Training details.}
We choose ResNet \cite{he2016} with 16 residual blocks and 32 channels as the backbone CNN model and append a pixel-wise sigmoid layer to output the probability of action $1$.
The model is trained for 200,000 iterations on the remaining 13,758 images with the Adam optimizer \cite{kingma2015}.
As halftones mainly depend on the input tone, we augment training images by jittering the brightness randomly (\textit{ColorJitter(brightness=0.9)} in torchvision).
The batch size is set as 64 and the learning rate is adjusted from 3e-4 to 1e-5 with the cosine annealing schedule \cite{loshchilov2017}.

\begin{figure}[t]
  \begin{minipage}[b]{0.32\linewidth}
    \centering
    \begin{overpic}[width=\linewidth]{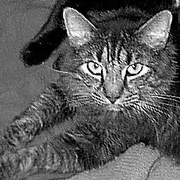}
      \put(83, 4){\footnotesize\colorbox{white}{\textbf{(a)}}}
    \end{overpic}
  \end{minipage}
  \hfill
  \begin{minipage}[b]{0.32\linewidth}
    \centering
    \begin{overpic}[width=\linewidth]{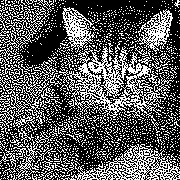}
      \put(83, 4){\footnotesize\colorbox{white}{\textbf{(b)}}}
    \end{overpic}
  \end{minipage}
  \hfill
  \begin{minipage}[b]{0.32\linewidth}
    \centering
    \begin{overpic}[width=\linewidth]{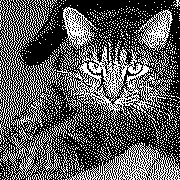}
      \put(83, 4){\footnotesize\colorbox{white}{\textbf{(c)}}}
    \end{overpic}
  \end{minipage}
  \vfill
  \vspace{1mm}
  \begin{minipage}[b]{0.32\linewidth}
    \centering
    \begin{overpic}[width=\linewidth]{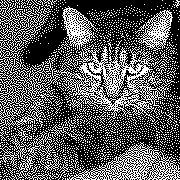}
      \put(83, 4){\footnotesize\colorbox{white}{\textbf{(d)}}}
    \end{overpic}
  \end{minipage}
  \hfill
  \begin{minipage}[b]{0.32\linewidth}
    \centering
    \begin{overpic}[width=\linewidth]{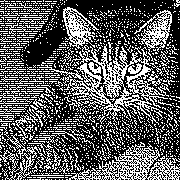}
      \put(83, 4){\footnotesize\colorbox{white}{\textbf{(e)}}}
    \end{overpic}
  \end{minipage}
  \hfill
  \begin{minipage}[b]{0.32\linewidth}
    \centering
    \begin{overpic}[width=\linewidth]{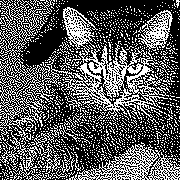}
      \put(83, 4){\footnotesize\colorbox{white}{\textbf{(f)}}}
    \end{overpic}
  \end{minipage}
  \vfill
  \vspace{-4mm}
  \caption{Cat halftone samples from different methods. (a) Continuous tone; (b) VAC(size=64); (c) Ostromoukhov; (d) DBS; (e) Xia et al.; (f) HALFTONERS.}
  \label{fig3}
\end{figure}

\begin{figure}[t]
  \begin{minipage}[b]{\linewidth}
    \centering
    \begin{overpic}[width=\linewidth]{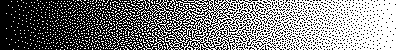}
      \put(95, 1){\footnotesize\colorbox{white}{\textbf{(a)}}}
    \end{overpic}
  \end{minipage}
  \vfill
  \vspace{0.5mm}
  \begin{minipage}[b]{\linewidth}
    \centering
    \begin{overpic}[width=\linewidth]{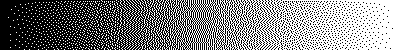}
      \put(95, 1){\footnotesize\colorbox{white}{\textbf{(b)}}}
    \end{overpic}
  \end{minipage}
  \vfill
  \vspace{0.5mm}
  \begin{minipage}[b]{\linewidth}
    \centering
    \begin{overpic}[width=\linewidth]{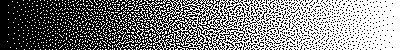}
      \put(95, 1){\footnotesize\colorbox{white}{\textbf{(c)}}}
    \end{overpic}
  \end{minipage}
  \vfill
  \vspace{0.5mm}
  \begin{minipage}[b]{\linewidth}
    \centering
    \begin{overpic}[width=\linewidth]{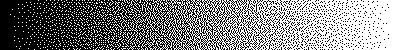}
      \put(95, 1){\footnotesize\colorbox{white}{\textbf{(d)}}}
    \end{overpic}
  \end{minipage}
  \vfill
  \vspace{0.5mm}
  \begin{minipage}[b]{\linewidth}
    \centering
    \begin{overpic}[width=\linewidth]{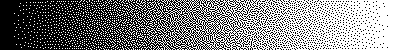}
      \put(95, 1){\footnotesize\colorbox{white}{\textbf{(e)}}}
    \end{overpic}
  \end{minipage}
  \vfill
  \vspace{-4mm}
  \caption{Gray ramp halftone samples. (a) VAC(size=64); (b) Ostromoukhov; (c) DBS; (d) Xia et al.; (e) HALFTONERS.}
  \vspace{-3mm}
  \label{fig4}
\end{figure}

\noindent
\textbf{Visual Comparison.}
Figure~\ref{fig3} shows halftone samples made by our method and other reference works, whereas Figure~\ref{fig4} shows the gray ramp halftone examples.
The structural details are preserved well in both Xia et al.'s and our results, while others suffer in the edge area.
Ostromoukhov's and Xia et al.'s methods introduce annoying artifacts in the flat area, which indicate bad blue-noise property, and it is tackled by the anisotropy suppressing loss function in our results.

\noindent
\textbf{Quantitative Comparison.}
The tone consistency is measured by PSNR between the HVS-filtered halftone and the HVS-filtered input continuous-tone.
The structural similarity is measured by SSIM between the halftone and the continuous-tone.
Table~\ref{table} shows the results.
We can see that our method achieves a competitive PSNR score and the best SSIM score.
According to the preference setting, one can trade SSIM for better PSNR by decreasing $\omega_s$ and vice versa.

\begin{figure}[t]
  \centering
  \includegraphics[width=\linewidth]{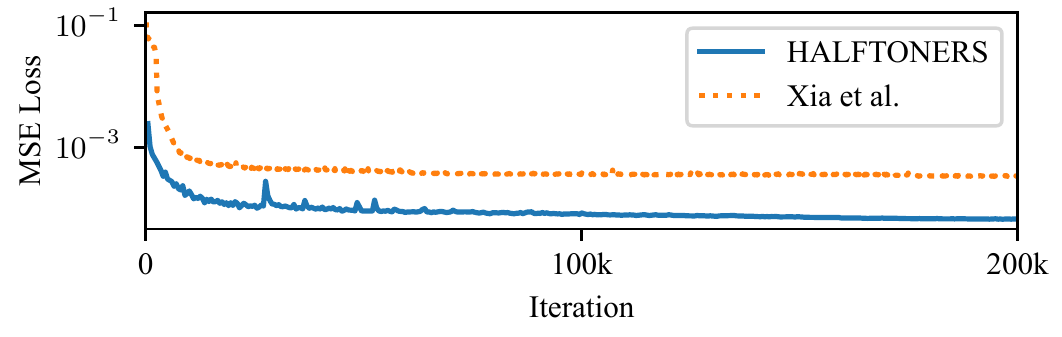}
  \vspace{-9mm}
  \caption{MSE training loss curves of Xia et al.'s and ours.}
  \label{fig5}
  \vspace{-5mm}
\end{figure}

\begin{table}[t]
  \centering
  \caption{Quantitative comparison with prior halftoning methods. ``*'' indicates this method is GPU-accelerated.}
  \label{table}
  \medskip
  \begin{small}
    \begin{tabular}{lcccc}
      \hline
      Method                               & PSNR  & SSIM  & Param. \# & Time (ms) \\
      \hline\hline
      VAC (64) \cite{ulichney1993}         & 29.19 & 0.082 & 4,096     & 0.04*     \\
      VAC (512) \cite{ulichney1993}        & 29.29 & 0.082 & 262,144   & 0.04*     \\
      Ostromoukhov \cite{ostromoukhov2001} & 33.07 & 0.102 & 384       & 18.64     \\
      DBS \cite{analoui1992}               & 34.00 & 0.084 & -         & 3008.71   \\
      Xia et al. \cite{xia2021}            & 29.31 & 0.159 & 37.8M     & 280.53*   \\
      \hline
      \textbf{Ours}                        & 31.32 & 0.176 & 298,945   & 171.50*   \\
      \hline
    \end{tabular}
  \end{small}
  \vspace{-3mm}
\end{table}

\noindent
\textbf{Learning Curves.}
We compare the MSE training loss curves of \cite{xia2021} (train the first stage to convergence) and ours (adopting Gaussian HVS for fair comparison).
As shown in Figure~\ref{fig5}, our method converges faster and achieves better results.

\noindent
\textbf{Processing Speed.}
We evaluate the processing time of all methods on a 1200x1200 image (shown in Table~\ref{table}).
Our method outperforms \cite{xia2021} in terms of image quality while being 1.6x faster with only 0.79\% parameters.
It is possible to close the speed gap between our model and error diffusion approaches by designing a more efficient CNN or using compression techniques \cite{deng2020}, and we leave it for future work.

\section{Conclusion}
\label{sec:conclusion}
In this work, we introduce HALFTONERS, a halftoning method based on multi-agent deep reinforcement learning.
Blue-noise halftone results with high quality are generated thanks to the low-variance policy gradient and the anisotropy suppressing loss function.
Additionally, our training procedure is more stable and simpler compared to existing deep halftoning methods.
In the meantime, the high parallelism of CNN makes it possible to deploy our model on real-time devices.
Promising future directions include designing more efficient NN architectures, constructing more complex HVS or device models, and expanding to multi-toning.

\vspace{-1mm}
~\\
\textbf{Acknowledgments.} This work is supported by the National Key R\&D Program of China (2021YFB2206200).

\vfill\pagebreak

\bibliographystyle{IEEEbib}
\bibliography{refs}

\end{document}